\def\year{2021}\relax

\documentclass[letterpaper]{article} % DO NOT CHANGE THIS
\def\mX{{\mathcal X}}
\def\mY{{\mathcal Y}}
\def\mZ{{\mathcal{Z}}}

\def\0{{\bf 0}}
\def\1{{\bf 1}}

\def\bx{{\bf x}}
\def\by{{\bf y}}

\def\citep{\cite}
\def\citet{\cite}

\newtheorem{prop}{Proposition}
\usepackage{aaai21}  % DO NOT CHANGE THIS
\usepackage{times}  % DO NOT CHANGE THIS
\usepackage{helvet} % DO NOT CHANGE THIS
\usepackage{amsmath}
\usepackage{array}
\usepackage{multirow}
\usepackage{amssymb}
\usepackage{graphicx}
\usepackage{courier}  % DO NOT CHANGE THIS
\usepackage[hyphens]{url}  % DO NOT CHANGE THIS
\usepackage{subfigure} % DO NOT CHANGE THIS
\usepackage{natbib}  % DO NOT CHANGE THIS AND DO NOT ADD ANY OPTIONS TO IT
\usepackage{caption} % DO NOT CHANGE THIS AND DO NOT ADD ANY OPTIONS TO IT
\title{Generative Model without Prior Distribution Matching} %Adversarial Embedding based Autoencoder }
\author{
    Cong Geng,  
   Jia Wang,
    Li Chen,
    Zhiyong Gao\\
}
\affiliations{
Shanghai Jiao Tong University \\
	Institute of Image Communication and Network Engineering \\
	Shanghai, China\\
   $ \mathrm{\{gengcong,jiawang,hilichen,zhiyong.gao\}@sjtu.edu.cn}$
}
\begin{document}

\maketitle

\begin{abstract}
Variational Autoencoder~(VAE) and its variations are classic generative models by learning a low-dimensional latent representation to satisfy some prior distribution (\textit{e.g.}, Gaussian distribution). Their advantages over GAN are that they can simultaneously generate high dimensional data and learn latent representations to reconstruct the inputs. However, it has been observed that a trade-off exists between reconstruction and generation since matching prior distribution may destroy the geometric structure of data manifold. 
To mitigate this problem, we propose to let the prior match the embedding distribution rather than imposing the latent variables to fit the prior. The embedding distribution is trained using a simple regularized autoencoder architecture which preserves the geometric structure to the maximum. Then an adversarial strategy is employed to achieve a latent mapping. We provide both theoretical and experimental support for the effectiveness of our method, which alleviates the contradiction between topological properties' preserving of data manifold and distribution matching in latent space. 
%In particular, our method is easy to converge as the prior only match the %distribution of low dimensional embeddings.
\end{abstract}

\section{Introduction}
Generative models represent complex data distributions using a generator function that maps low-dimensional latent vectors subjected to a specified distribution to high-dimensional data outputs. Variational autoencoder (VAE)~\cite{kingma_auto-encoding_2014} and generative adversarial network~(GAN)~\cite{Goodfellow2014Generative} are two notable deep learning generative models. Compared with GAN, VAE has the benefit of an
encoder that learns latent representation of data inputs, making VAE an effective tool for generating and understanding manifold structure in high-dimensional data.
Traditional VAE and its variants are trained by minimizing a reconstruction error and a divergence to force the variational posterior to fit the prior. However, it's hard for the latent embedding to simultaneously keep the topological properties of the data manifold and satisfy a Gaussian distribution. It means the performance of VAEs is a trade-off between reconstruction and generation.

In this work, we present a generative model without prior distribution matching to solve the above trade-off. Different from VAEs which encourage the approximate posterior to match the prior, we propose a latent mapping letting the prior fit the embedding distribution.
The key point of our method is the learning of the embedding distribution which is expected to preserve the structure of data manifold and is easy for the prior to learn. David Berthelot et al.~\cite{berthelot2019Understandinga} propose a regularization procedure that encourages interpolated outputs to appear more realistic using an adversarial learning strategy. It improves representation learning performance on downstream tasks. We extend this method to our embedding learning task and find it can help to capture the characteristics of embedding distribution. To avoid the learned embedding distribution to be over-dispersed, we introduce a batch normalization trick on the latent space based on the volume concentration of high dimensional sphere. After learning a useful latent representation, we employ a GAN structure to encourage the prior to match the embedding distribution. With our method, we can generate high dimensional data sampled from a specified prior while reconstructing observations from latent representations which preserve the topological properties of data manifolds. Overall, the main contributions of this work are as follows:

\begin{itemize}
\item We explain the causes of the existing trade-off in most variational based autoencoders and provide a theoretical analysis of our method on alleviating this trade-off problem.
\item We introduce an autoencoder structure with an adversarial interpolation regularization and a batch normalization trick to obtain a learning-facilitated latent embedding representation that preserves the topological structure of the data manifold. 
\item We propose an adversarial based latent mapping in order to sample from a specified prior distribution when generating new high-dimensional data. 
\end{itemize}
% Due to the aforementioned strategies, we can reconstruct data manifold using a latent embedding representation and simultaneously generating new data sampling from a prior distribution rather than getting a trade-off between reconstruction and generation.

\section{Related Work}
Autoencoder (AE) networks~\cite{2010Stacked} are unsupervised approaches aiming at combining the “reconstruction” as well as the “representation” properties by learning an encoder-generator map. Since then, a lot of progress has been made based on autoencoders. These variants are generally divided into two categories: VAE-based and GAN-based. In the VAE-based case, AEs are regularized to explicitly match the distribution of the latent representations with a predefined prior. In the VAE-based case, VAE~\cite{kingma_auto-encoding_2014} introduces an additional loss into the reconstruction loss. This loss measures the KL divergence between the distribution of the latent representations and the prior. Along with learning
continuous representations, vector-quantized VAE (VQVAE)~\cite{2017Neural} uses a vector quantization method allowing the model to circumvent issues of “posterior collapse. In the GAN-based case, adversarial AE (AAE)~\cite{2015Adversarial} introduces an auxiliary subnetwork referred to as a discriminator based on the GAN framework and forces this
discriminator to measure the difference between latent representations and the prior. Besides, Wasserstein autoencoder(WAE)~\cite{Tolstikhin2017Wasserstein} also employs MMD to measure the distribution divergence. Moreover, VAE-GAN~\cite{vaegan} fuses both the VAE
and GAN frameworks in the autoencoder which replaces the traditional pixel-wise reconstruction loss by an adversarial feature-wise reconstruction loss obtained from the GAN’s discriminator. Another famous integration for VAE and GAN is introVAE~\cite{2018IntroVAE}. It requires no extra discriminators because the inference model itself serves as a discriminator to distinguish between the generated and real samples. 

For latent representation, ACAI~\cite{berthelot2019Understandinga} proposes a regularization procedure which encourages interpolated outputs to appear more realistic by fooling a critic network which has been trained to recover the mixing coefficient from interpolated data. It demonstrates good interpolation can result in a useful representation which is more effective on downstream tasks. MI-AE~\cite{2019Improving} improves ACAI by introducing a multidimensional interpolation approach for each dimension of the latent representations and encouraging generated data points to be realistic in the GAN framework. Besides, S-VAE~\cite{davidson2018hyperspherical,2018Spherical} and SAE~\cite{zhao2019latent} project latent variable on a sphere to improve learning on high-dimensional space. Other autoencoder techniques provide frameworks that attempt to shape the latent space with respect to factor disentanglement~\cite{Bouchacourt2017Multi, disentanglement}.

\section{Problem Definition}
\paragraph{Notation.}
Throughout the paper, we use the following notations.
We use calligraphic letters (\textit{e.g.}, $ \mX $) to denote spaces, and bold lower case letters (\textit{e.g.}, $ \bx $) to denote vectors.  %bold capital letters (\textit{e.g.}, $ \bX $) to denote the matrix, and
% 	Let $\sgn(\cdot)$ be the sign function.
Let $p(x)$ be the real data distribution, and
let $ P_n {=} \{\bx_i, \by_i\}_{i=1}^n {\in} \mX {\times}  \mY $ be the training data, where $\mX $ denotes the data space.

% Most existing AE-based generative models (\textit{e.g.}, VAE, AAE and WAE) enforce the distribution of embeddings to be close to some prior distribution (\textit{e.g.}, Gaussian distribution or Uniform distribution). 
% Specifically, the optimization problem can be written as:
% \begin{align}
%     \mathbb{E}_{x}[\| g(\phi(x)){-} x\|] + d (q_\phi(z|x), p(z)),
% \end{align}
% where $p(z)$ is a prior distribution and $d(\cdot, \cdot)$ is some distribution divergence.
% The following example illustrates that the KL divergence and the reconstruction loss are hard to be guaranteed.

\begin{figure}[t]
	\centering
	\includegraphics[width=0.9\linewidth]{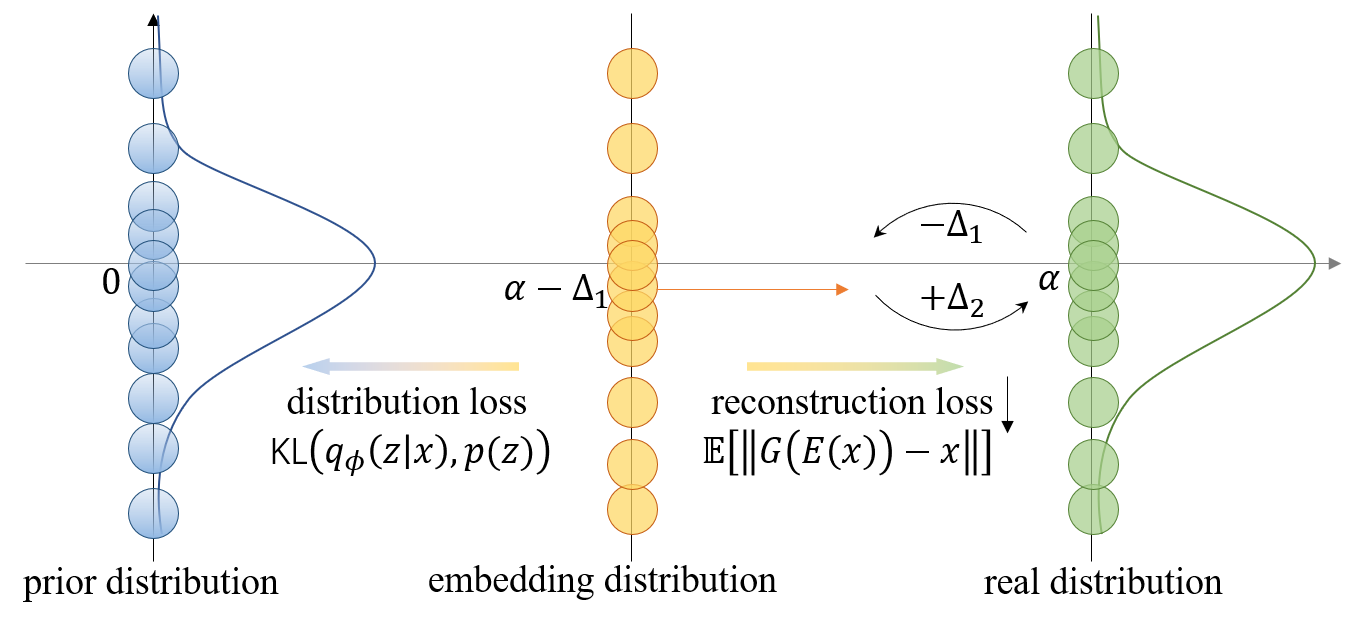} 
	\caption{Intuitive understanding of Proposition \ref{prop:2}}
	\label{fig:prop}
\end{figure}

\subsection{Trade off in Variational Autoencoder}
Traditional VAE models the distribution of observations $p_{\theta}(x)$ by specifying a prior $p(z)$ along with a likelihood $p_{\theta}(x|z)$ that connects it with the observation:
\begin{equation}
	\label{1}
	p_{\theta}(x)=\int{p_{\theta}(x|z)p(z)dz}
\end{equation}
The integral for computing $p_{\theta}(x)$ is intractable, making it hard to maximize the marginal likelihood of the model under the data. To overcome this intractability, VAEs instead maximize the Evidence Lower Bound (ELBO) of the marginal likelihood:
\begin{equation}
	\log p_{\theta}(x)\geq \mathbb{E}_{q_\phi(z|x)}[\log p_{\theta}(x|z)]-KL[q_\phi(z|x)||p(z)],
\end{equation}
where $q_\phi(z|x)$ is the variational posterior. The neural networks used to parameterize $q_\phi(z|x)$ and $p_\theta(x|z)$ are referred to as the encoder and decoder, respectively. We can observe that the second term in ELBO, the Kullback-Leibler (KL) divergence captures how distinct the conditional distribution of latent representation corresponding to each training example is from the prior $p(z)$. The VAE objective minimizes KL divergence to force the encoder to output $\mu(x_i) = 0$ and $\sigma(x_i) = 1$ for all samples. In this case, the decoder will face an impossible task of reconstructing different samples from completely random noise which is called "posterior collapse". The following case will illustrate this phenomenon.
\begin{prop} \label{prop:vae}
Assume $p(z)$ is some specified prior distribution. $p(x)$ is the real data distribution, if we let $KL[q_\phi(z|x)||p(z)]=0$, for every $x \in \mX $, then ELBO is globally
optimized only if $p_\theta(x|z)= p(x)$ for every $z \in \mZ $.
\end{prop}
Proposition \ref{prop:vae} means VAE will completely lose reconstruction ability if we force $KL[q_\phi(z|x)||p(z)]=0$.
To avoid this,  VAE requires manually fine-tuning the weight of the KL component and reconstruction hyper-parameters. In order to realize reconstruction, the variational posterior $q_\phi(z|x_i)$ is a Gaussian distribution whose mean and variance are related to the encoder and training sample $x_i$. That is,
\begin{equation}
	q_\phi(z|x_i)\thicksim \mathcal{N}\big(\mu(x_i),\sigma(x_i)\big)
\end{equation}
which is not guaranteeing that the overall encoded distribution $\mathbb{E}_{p(x)} [q_\phi(z|x)]$ matches $p(z)$. 
Figure~\ref{encodeing_results}~(a) shows the latent embedding trained by VAE. It illustrates this mismatching. The following example provides an extreme case to show that the KL divergence term is hard to be guaranteed if we incline to the reconstruction loss.
 \begin{prop} \label{prop:2}
Suppose $\mX, \mZ$ are 1-D spaces. Assume $p(z)$ is some specified continuous prior distribution, $p(x)$ is a 1-D continuous distribution with finite support, $q_\phi(z|x)$ is continuous in finite support region for different $x$. then 
\begin{equation}
       \sup_x KL[q_\phi(z|x)||p(z)]=+\infty,
\end{equation}
when $\mathbb{E}_{p(x)}\big[\mathbb{E}_{q_\phi(z|x)}[log p_{\theta}(x|z)]\big]$ achieves the maximum.
\end{prop}
Figure.~\ref{fig:prop} intuitively explains Proposition~\ref{prop:2}, which reveals that the KL divergence and the reconstruction loss may be hard to be guaranteed to be small simultaneously.
To alleviate this problem, infoVAE~\cite{zhao2019infovae} introduces a mutual information term into original ELBO to weight the preference between correct inference and fitting the data distribution, and specify a preference on how much the model should rely on the latent variables:
\begin{equation}
\begin{aligned}
\mathcal{L}_{\text {InfoVAE }}=&-\lambda KL[q_{\phi}(z) \| p(z)]- \\
\quad & \mathbb{E}_{q(z)}\big[KL[q_{\phi}(x | z) \| p_{\theta}(x | z)]\big]+\alpha I_{q}(x ; z) \\
 \equiv & \mathbb{E}_{p(x)} \mathbb{E}_{q_{\phi}(z \mid x)}\left[\log p_{\theta}(x|z)\right]-\\
&(1-\alpha) \mathbb{E}_{p(x)} \big[KL[q_{\phi}(z \mid x) \| p(z)]\big]-\\
&(\alpha+\lambda-1) KL[q_{\phi}(z) \| p(z)]
\end{aligned}
\end{equation}
Essentially, infoVAE weights the KL term in original ELBO and adds another KL term $ KL[q_{\phi}(z) \| p(z)]$ to further alleviate the contradiction between reconstruction and generation.
When $\alpha=1, \lambda=1$, infoVAE can degenerate into AAE~\cite{2015Adversarial} which is equivalent to adding a mutual information term into the ELBO bound.
WAE~\cite{Tolstikhin2017Wasserstein} minimizes the optimal transport cost $W_c\big(P_X;P_\theta(X)\big)$ based on the novel autoencoder formulation, leading to the following objective:
% It can be divided into two steps: the decoder tries to reconstruct the encoded training examples as measured by the cost function $c$. The encoder tries to match the encoded distribution of training examples $q_\phi(z) := \mathbb{E}_{p(x)}[q_\phi(z|x)]$ to the prior $p(z)$ as measured by any specified divergence $D_Z\big(Q_\phi(Z),P_Z\big)$, 
 
\begin{equation}
	\label{wae}
	\begin{aligned}
		D_{WAE}\big(P_X;P_\theta(X)\big)&=\inf_{Q_\phi(Z)}\mathbb{E}_{p(x)}\mathbb{E}_{q_\phi(z|x)}\big[c\big(x,G(z)\big)\big]\\ &+\lambda \cdot D_Z\big(Q_\phi(Z),P_Z\big)
	\end{aligned}
\end{equation}
WAE can be viewed as another version of AAE from the Wasserstein distance perspective.
However, all these variants try to simultaneously achieve two conflicting goals: preserving the topological properties of data manifold and making sure that the latent codes provided to the decoder are informative enough to generate meaningful samples. 
% Fortunately, the optimal transport cost takes a simple form of discrepancy measures between data distribution $P_X$ and model $P_\theta(X)$ through the autoencoder:
% \begin{equation}
% 	\label{wc}
% 	W_c\big(P_X;P_\theta(X)\big)=\inf_{Q_\phi(Z)=P_Z}\mathbb{E}_{p(x)}\mathbb{E}_{q_\phi(z|x)}\big[c\big(x,G(z)\big)\big],
% \end{equation}
% where $G$ is the decoder of the autoencoder architecture and $W_c$ is the wasserstein distance with cost function $c$~\cite{villani2008optimal}. The proof of equation~\ref{wc} can be found in~\cite{bousquet2017optimal}. One way to implement
% a numerical solution is to relax the constraints on $Q_\phi(Z)$ by adding a penalty to the objective. This leads to the WAE objective:
In the actual training process, such autoencoders prone to get a trade-off between reconstruction and generation. The conflicting phenomenon is depicted in Fig.~\ref{encodeing_results}~(b).
\section{Proposed Method}
To solve the above problems, we expect to separate the two tasks of reconstruction and generation. Instead of imposing the latent representation match the prior, we let the prior fit the latent embedding distribution. 
For the purposes of optimization, we rewrite the reconstruction term in ELBO as follows:
\begin{equation}
	\begin{aligned}
		&\mathbb{E}_{p(x)} \mathbb{E}_{q_{\phi}(z | x)}\left[\log p_{\theta}(x|z)\right]= \int q_{\phi}(z, x)\log p_\theta(x|z)dzdx\\
		%&=\int q_{\phi}(z, x)log\frac{p_\theta(x,z)}{p(z)}dzdx\\
		&=\int q_{\phi}(z, x) \log p_\theta(x,z)dzdx - \int q_{\phi}(z, x) \log p(z)dzdx\\
	&	=\int q_{\phi}(z, x) \log p_\theta(x,z)dzdx - \int q_{\phi}(z) \log p(z)dz
	\end{aligned}
\label{1}
\end{equation}
 Equation~\ref{1} can be transposed to:
 \begin{equation}
 	\begin{aligned}
 		\int q_{\phi}(z, x) \log p_\theta(x,z)dzdx= \mathbb{E}_{p(x)} \mathbb{E}_{q_{\phi}(z | x)}\left[\log p_{\theta}(x|z)\right]
 		\\ + \int q_{\phi}(z) \log p(z)dz
\end{aligned}
\label{2}
\end{equation}
If we fix our encoder, equation~\ref{2} can obtain the maximum value when $q_{\phi}(z, x)=p_\theta(x,z)$. At this point, the two terms on the right respectively achieve the maximum. We need to impose $p_{\theta}(x|z)=q_{\phi}(x|z)$  and $p(z)=q_{\phi}(z)$. For a simple autoencoder, if we fix the encoder, it's flexible enough to find a decoder to satisfy $p_{\theta}(x|z)=q_{\phi}(x|z)$. Then we only need to make the prior match the embedding distribution, which is also not hard to achieve. There exists no trade-off in our framework between generation and reconstruction. 

\begin{figure*}[t]
	\centering
	\includegraphics[width=0.9\linewidth]{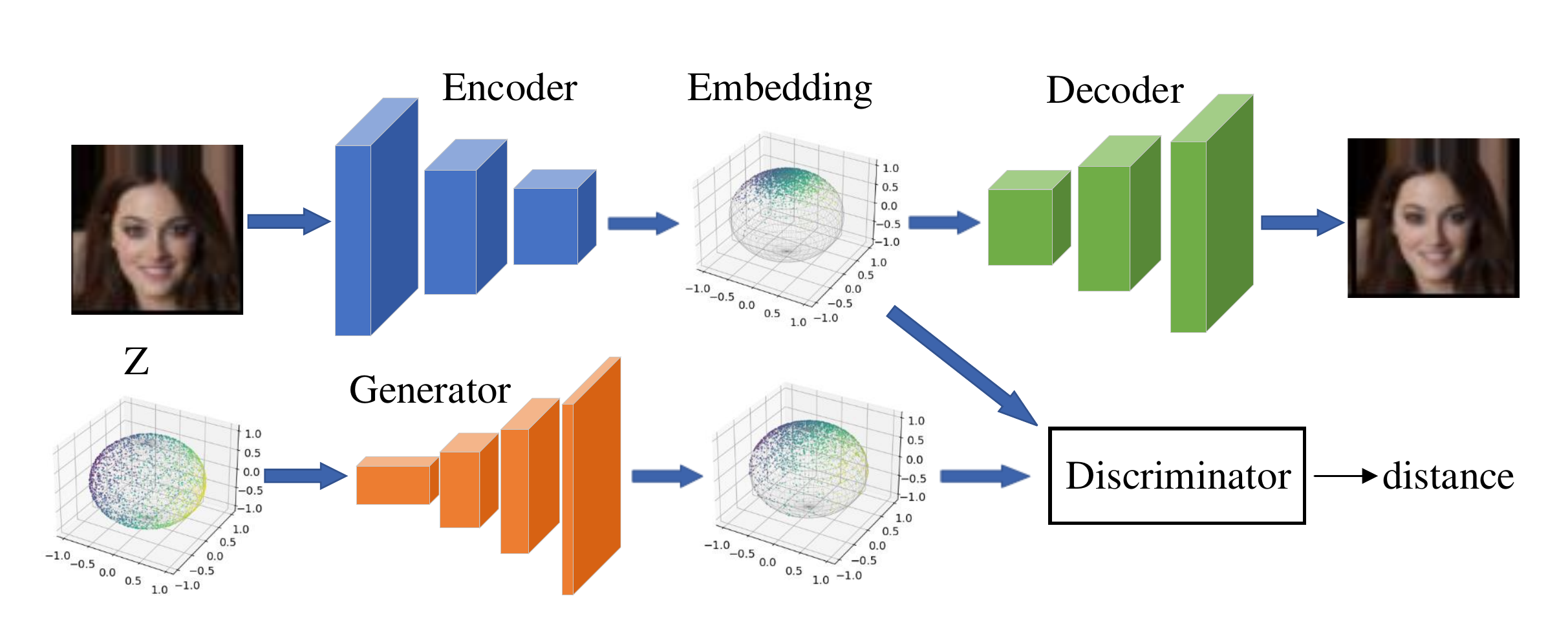} % Reduce the figure size so that it is slightly narrower than the column. Don't use precise values for figure width.This setup will avoid overfull boxes.
	\caption{Overall Framework}
	\label{fig2}
	\vspace{-10pt}
\end{figure*}

\begin{figure}[htpb]
	\footnotesize
	\centering
	\renewcommand{\tabcolsep}{1pt} \renewcommand{\arraystretch}{0.9} \begin{tabular}{cc}
		\includegraphics[width=0.5\linewidth]{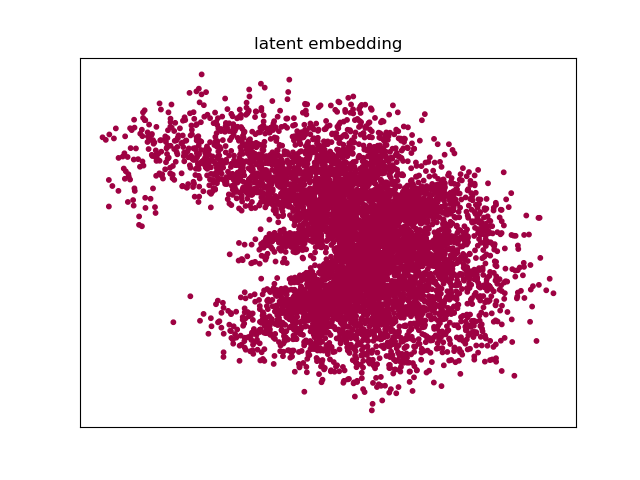} &
		\includegraphics[width=0.5\linewidth]{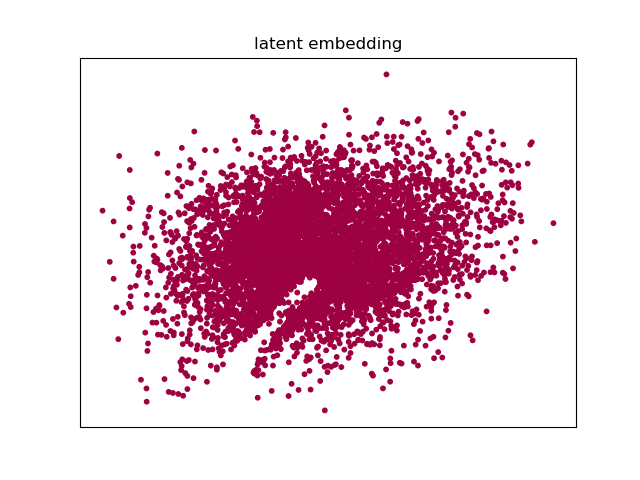}
		\\
		(a) VAE & (b) AAE \\
	\end{tabular}
	\vspace{-5pt}
	\caption{
		The encoding results with different methods for digit "0" in MNIST Dataset.
	}
	\label{encodeing_results} \vspace{-10pt}
\end{figure}

We introduce an extra generator $g$ to match the latent embedding distribution from the prior. Then the decoder $G$ maps latent samples $g(z)$ to data space $x=G(g(z))$. 
% We denote the distribution of $G(g(z))$ as $P_{Gg}$ and the distribution of $g(z)$ as $P_g$. The Wasserstein distance between $P_X$ and $P_{Gg}$ can be derived according to Equation~\ref{wae}.
% \begin{equation}
% 	\label{wae2}
% 	\begin{aligned}
% 		D_{WAE}\big(P_X;P_{Gg}\big)&=\inf_{q_\phi(z|x)}\mathbb{E}_{p(x)}\mathbb{E}_{q_\phi(z|x)}\big[c\big(x,G(z)\big)\big]\\ &+\lambda \cdot D_Z\big(Q_\phi(Z),P_g\big)
% 	\end{aligned}
% \end{equation}
% The objective can be separated into two parts. The first term is a reconstruction loss to train the autoencoder. The second term is a divergence loss to train the generator $g$.
For implementation, we use the adversarial training to estimate the distance metric between distributions. Fig~\ref{fig2} shows our overall framework.
\subsection{Autoencoder with an Adversarial Regularizer}
The major challenge for our method is the determination of our latent representation. The latent space trained by autoencoder is expected to be concentrated and easy to learn for the prior. Therefore, inspired by ACAI~\cite{berthelot2019Understandinga} and MI-AE~\cite{2019Improving}, we add an adversarial regularizer to the original autoencoder loss to improve our latent representation. Similar to ACAI, we use a discriminator $d$ to form a GAN with our autoencoder to propel the linear interpolated data to perceptually approximate real data to as realistic an extent as possible. The objective of the discriminator can be reformulated as below:
\begin{equation}
	\begin{aligned}
		L_{dis}=&\|d(x)\|^2+\|d\big(\gamma x+(1-\gamma)\hat{x}\big)-\lambda\|^2+\\&\|d(x_\mu)-\mu-\lambda\|^2
	\end{aligned}
\end{equation}
where, as above, $\hat{x}$ is the reconstruction of $x$ through the autoencoder $\hat{x}=G(E(x))$, $x_\mu=G\big(\mu E(x_1)+(1-\mu) E(x_2)\big)$. $\lambda$ and $\gamma$ are two hyperparameters. $\mu$ is randomly sampled from
the uniform distribution on [0, 0.5]. Different from ACAI and MI-AE, we constrain the discriminator's output of the interpolation between $x$ and $\hat{x}$ to be the preset $\lambda$. Because we need to distinguish between the subtle difference between reconstruction data and real data to guide the encoder to obtain a better latent representation. Since the $x_\mu$ is less realistic than $\hat{x}$, we predict $\mu+\lambda$ for $x_\mu$. With this loss function for discriminator, the autoencoder’s objective is modified by adding two regularization term:
\begin{equation}
	L_{ae}=\|x-G(E(x))\|^2+\omega_1\|d(x_\mu)\|^2+\omega_2\|d(\hat{x})\|^2
\end{equation}
where $\omega_1, \omega_2$ are hyper-parameters for adjusting the weights of the above losses. 
\subsection{Latent Normalization}
In high dimensional space, VAE suffers from the dimensional dilemma that can be interpreted via some counterintuitive geometric facts. The data examples for a dataset become rather sparse in high dimensions since the geometric property reveals that the volume ratio between a cube and its inscribed sphere goes to infinity when the dimension goes very large~\cite{van2014probability}. Therefore, it becomes challenging to fit a distribution in high dimensions. Based on these surprising phenomenons, we employ a simple batch normalization trick to make our embedding subjected to a distribution with its mean close to zero and variance to 1. This operation makes embedding distribution more concentrated and is easier for the prior to learn.
The operation in the training process can be easily performed by:
\begin{equation}
	x \stackrel{E}\longmapsto z \longmapsto BN(z) \longmapsto \hat{z} \stackrel{G}\longmapsto \hat{x}
\end{equation}
In fact, this BN trick can be viewed as a spherical projection in high dimensional space since volume concentration~\cite{blum2016foundations} says that the volume of the sphere in the high-dimensional space is highly concentrated near the surface. The interior is nearly empty.
The benefit of spherical projection over other latent regularization is that the 2-Wasserstein distance between two arbitrary sets of random variables randomly drawn on the sphere converges to a constant when the dimension is sufficiently large. Furthermore, Deli Zhao et al~\cite{zhao2019latent} prove that the distance convergences to $\sqrt{2nr}$ where $n$ is the number of the latent variables in the dataset and $r$ is the sphere radius. This theorem illustrates the latent variables on the sphere are distribution-robust. However, It holds under a critical condition that latent vectors are randomly drawn from the sphere. Sometimes the condition violates if the prior and latent embedding are sampled from different orthants. Therefore, \cite{zhao2019latent} proposes a very simple approach by centralizing $z_i-\bar{z_i}\mathbf{1}$, where $z_i=[z_i^1,z_i^2,\cdots,z_i^{d_z}]$ and the mean $\bar{z_i}=\frac{1}{d_z}\sum_{j}z_i^j$. But this centralization will 
destroy the geometry of the latent embedding, which is against our original intention. Thanks to the BN operation, we can obtain a latent distribution with zero meanwhile preserving the geometric structure. After this latent normalization, we can guarantee that the distance between the prior and latent representation is small enough to alleviate the next latent mapping. 
\begin{figure}[t]
	\footnotesize
	\centering
	\renewcommand{\tabcolsep}{1pt} \renewcommand{\arraystretch}{0.9} \begin{tabular}{c}
		\includegraphics[width=\linewidth]{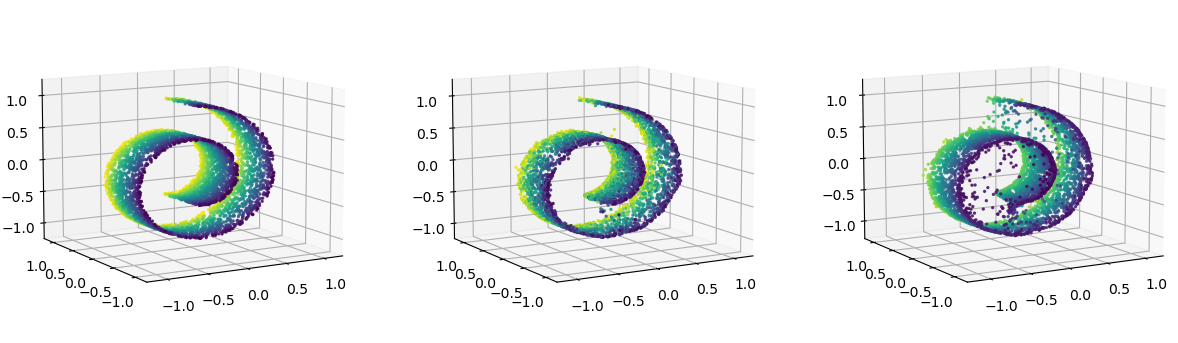} \\
	   (a) VAE\\
		\includegraphics[width=\linewidth]{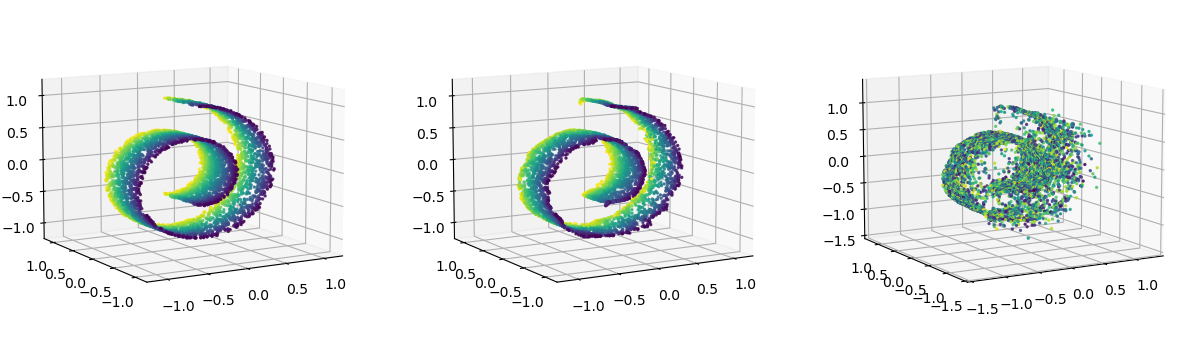}\\
	  (b) AAE \\
		\includegraphics[width=\linewidth]{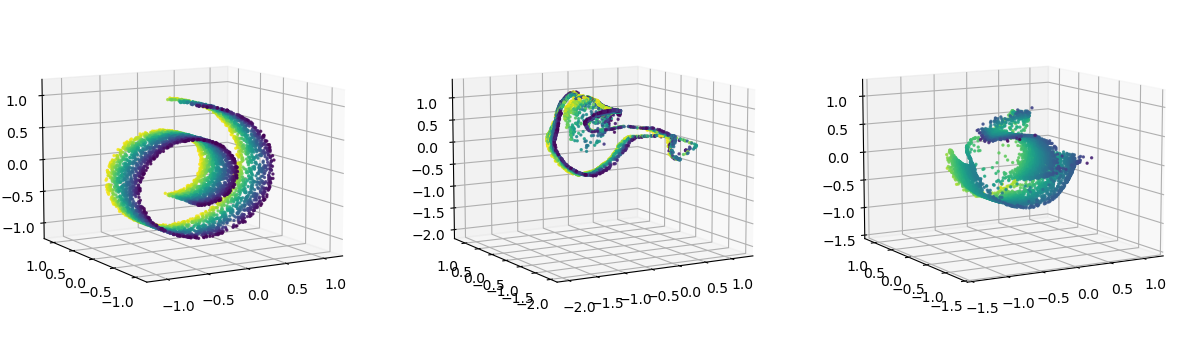}\\
		(c) ALAE \\
		\includegraphics[width=\linewidth]{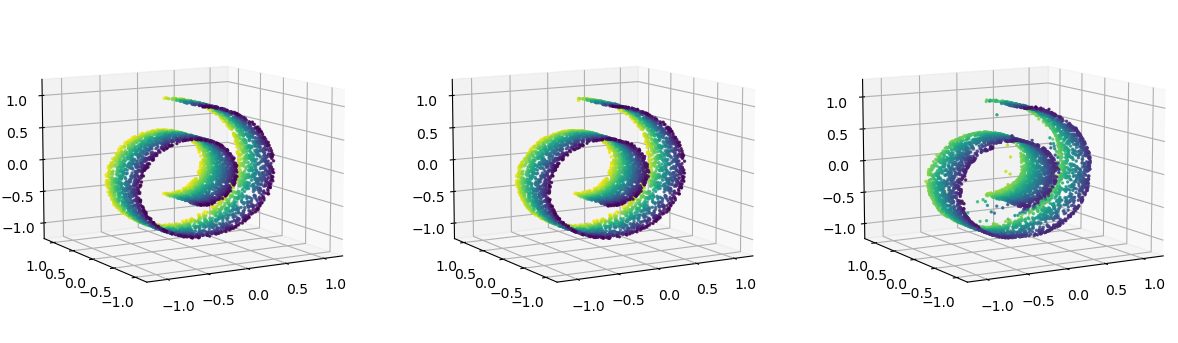}
		\\
		 (d) Ours \\
	\end{tabular}
	\vspace{-5pt}
	\caption{
		The reconstructed and generated data manifolds for swiss-roll. The left column is the input manifold, the middle column is the reconstruction manifold, and the right column is the generated manifold from Gaussian prior.
	}
	\label{recon_gene_swissroll} \vspace{-10pt}
\end{figure}

\begin{figure}[t]
	\subfigure[VAE]{
	\centering
		%\hspace{1cm}
		\includegraphics[width=0.45\linewidth]{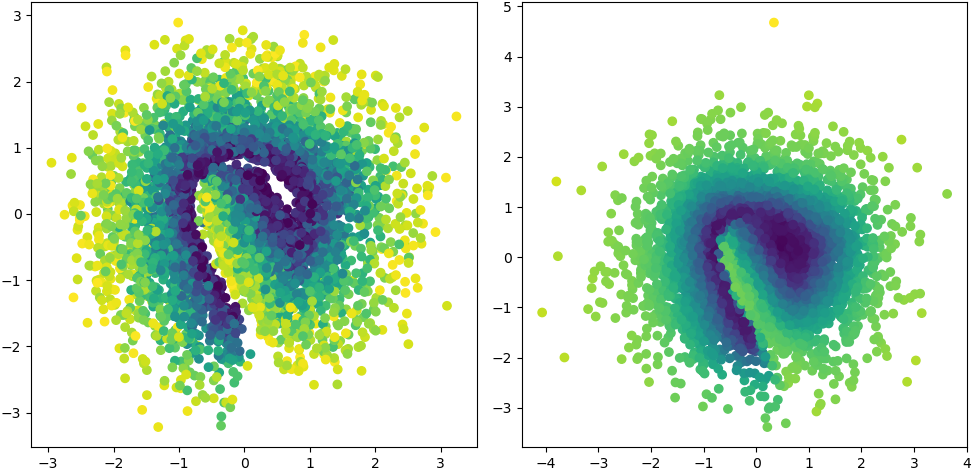}}
		\quad
	\subfigure[AAE]{
	\centering
	%\hspace{1cm}
		\includegraphics[width=0.45\linewidth]{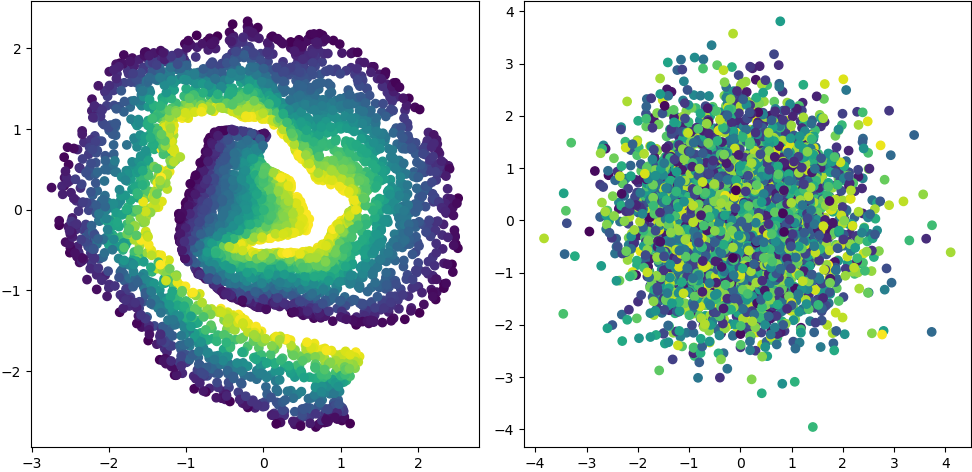}}
	
	\subfigure[ALAE]{
	\centering
		\includegraphics[width=0.97\linewidth]{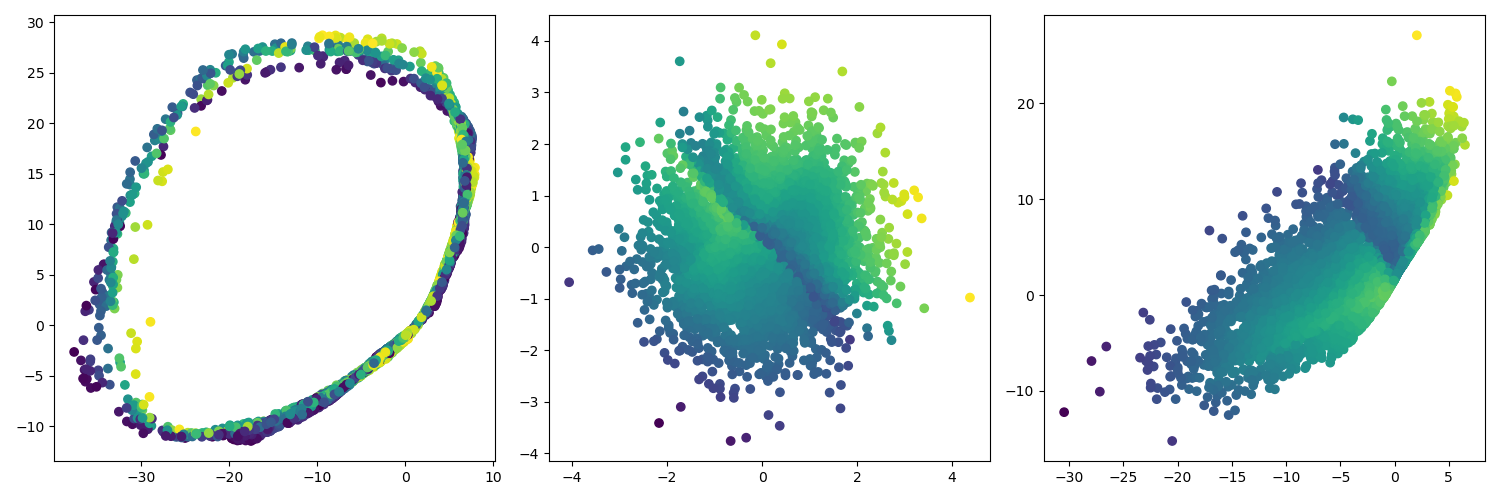}}
		
	\subfigure[Ours]{
	\centering
		\includegraphics[width=0.97\linewidth]{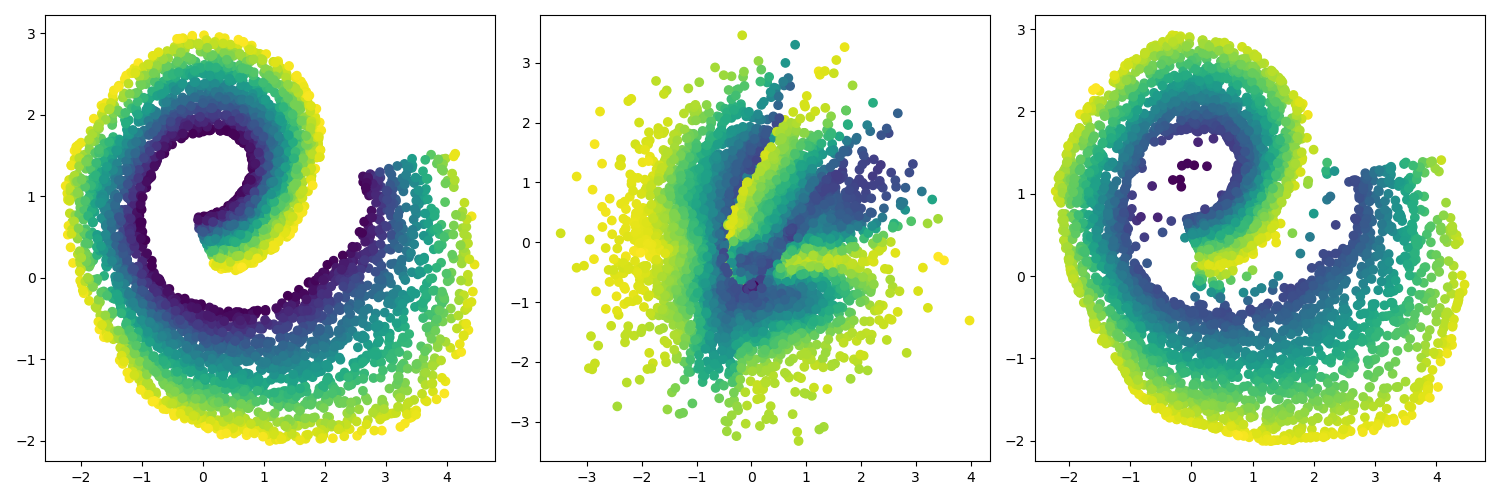}}
	%\hspace{0.15cm}
% 	\begin{subfigure}{0.5\textwidth}
		
% 		% column 3
% 		\begin{minipage}{\linewidth}
% 			\begin{minipage}{\linewidth}
% 				\hspace{-0.15cm}
% 				\includegraphics[width=0.815\linewidth,height=0.6cm,right]{swiss_roll/emb_ours.png}
% 			\end{minipage}
% 		\end{minipage}%     
		
% 		\vspace{-0.2cm}
% 		\caption{AAE}
% 	\end{subfigure}
%	\vspace{-0.2cm}
	\caption{Latent distributions for swiss-roll. The left image is the latent embedding, the middle image is the Gaussian prior. For ALAE and our method, the right image is the generated latent distribution from Gaussian prior.}
	\label{latent_representation}
\end{figure}
\subsection{Latent Mapping}
After getting the desired latent representation, we need to let the specified prior match this latent distribution. GAN based divergence estimation is widely applied in deep learning. Thus we introduce a GAN structure for our latent mapping. Specifically, we introduce a discriminator $D$ in the
latent space trying to separate $g(z)$ with $z$ sampled from $P_Z$ and latent embeddings $E(x)$ subject to $Q_\phi(Z)$. For network training, we employ the hinge version of the adversarial loss which were applied in SAGAN~\cite{zhang2019SelfAttention} and BigGAN~\cite{brock2019Large}, resulting in the following discriminator objective:
\begin{equation}
	\begin{aligned}
		L_D=&\mathbb{E}_{x\thicksim p_r(x)}[relu(1-D(E(x)))]+\\&\mathbb{E}_{z\thicksim p(z)}[relu(1+D(g(z)))]
	\end{aligned}
\end{equation} 
For generator $g$, the loss function can be written as follows:
\begin{equation}
	\begin{aligned}
		L_{g}=\mathbb{E}_{z\thicksim p(z)}\big[-D(g(z))]
		%+&\beta_1\|d(G(g(z)))\|^2+\\&\beta_2\|e(g(z))-z\|_1\big]
	\end{aligned}
\end{equation}

\begin{figure*}[t]
	\footnotesize
	\centering
	\renewcommand{\tabcolsep}{1pt}
	\renewcommand{\arraystretch}{0.8} 
	\begin{tabular}{ccccc}
         \multirow{1}{*}[40pt]{Without BN}&
		\includegraphics[width=0.2\linewidth]{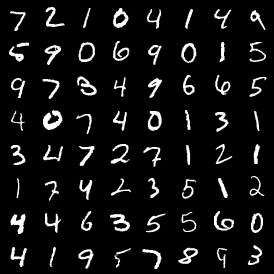} &
		\includegraphics[width=0.2\linewidth]{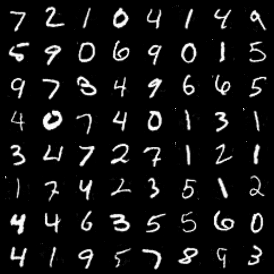}&
		\includegraphics[width=0.2\linewidth]{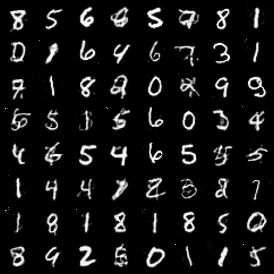}&
		\includegraphics[width=0.2\linewidth]{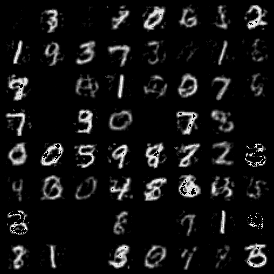}
		\\
	\multirow{1}{*}[40pt]{With BN}&
		\includegraphics[width=0.2\linewidth]{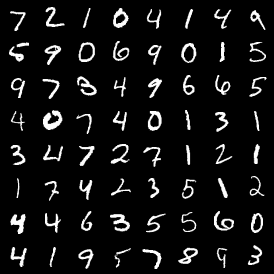} &
		\includegraphics[width=0.2\linewidth]{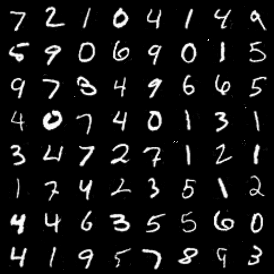}&
		\includegraphics[width=0.2\linewidth]{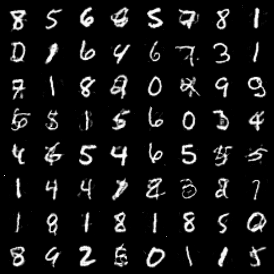}&
		\includegraphics[width=0.2\linewidth]{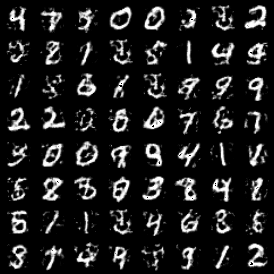}
		\\
		\multirow{1}{*}[40pt]{BN+Regularizer}&
		\includegraphics[width=0.2\linewidth]{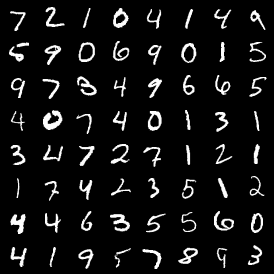} &
		\includegraphics[width=0.2\linewidth]{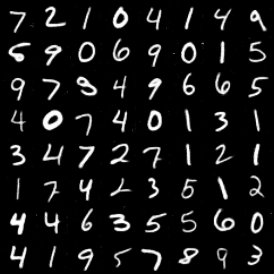}&
		\includegraphics[width=0.2\linewidth]{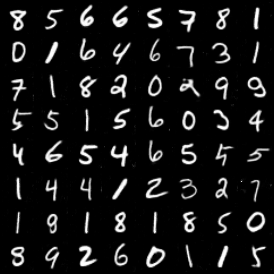}&
		\includegraphics[width=0.2\linewidth]{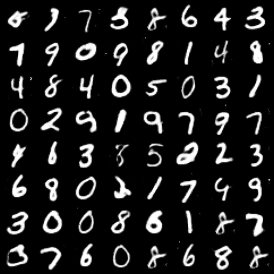}
		\\&(a) origin & (b) reconstruction &  (c) interpolation & (d) generation
	\end{tabular}
	\vspace{-5pt}
	\caption{Results on MNIST Dataset. BN represents batch normalization operation.
	}
	\label{recon_gene} \vspace{-10pt}
\end{figure*}
\begin{figure*}[b]
	\footnotesize
	\centering
	\renewcommand{\tabcolsep}{1pt} \renewcommand{\arraystretch}{0.9} \begin{tabular}{ccc}
		\includegraphics[width=0.3\linewidth]{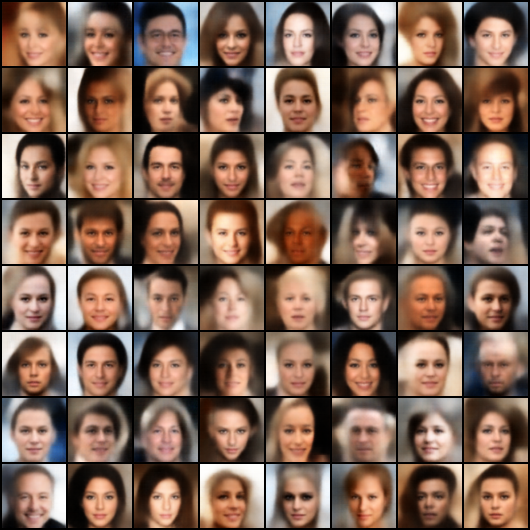} &
		\includegraphics[width=0.3\linewidth]{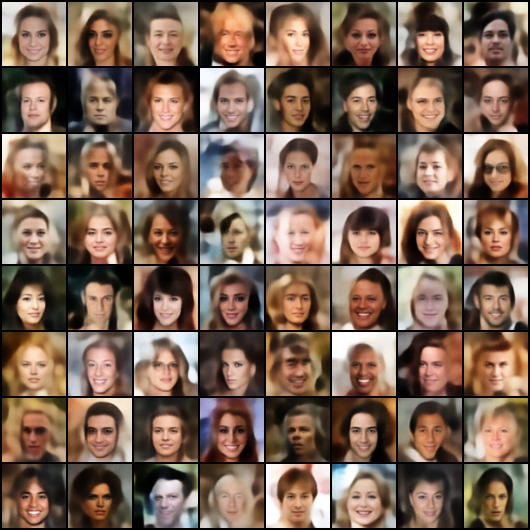} &
		\includegraphics[width=0.3\linewidth]{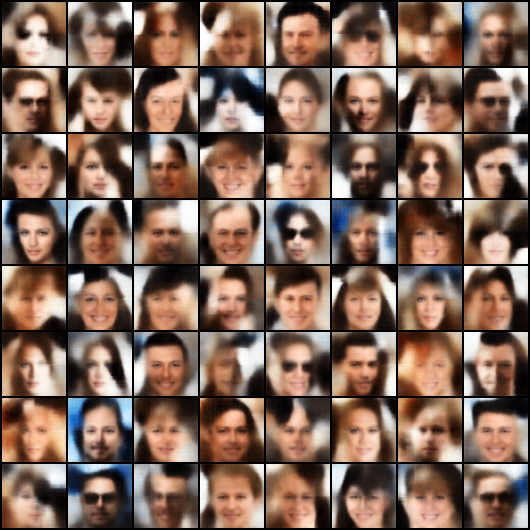}\\
		(a) VAE & (b) AAE & (c) WAE  \\
		\includegraphics[width=0.3\linewidth]{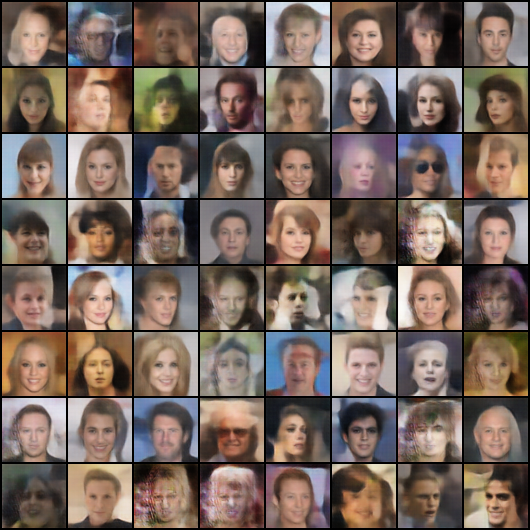} &
		\includegraphics[width=0.3\linewidth]{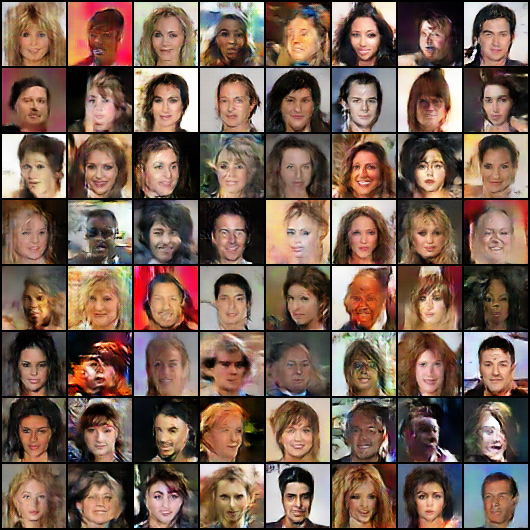} &
		\includegraphics[width=0.3\linewidth]{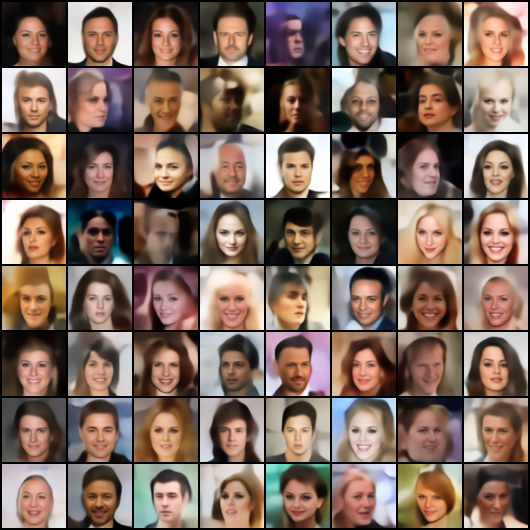}\\
		(d) IntroVAE & (e) AGE & (f) Ours
		\\
	\end{tabular}
	\vspace{-5pt}
	\caption{
		The generated results with different methods for CelebA Dataset.
	}
	\label{gene} \vspace{-10pt}
\end{figure*}
\begin{figure*}[b]
	\footnotesize
	\centering
	\renewcommand{\tabcolsep}{1pt} \renewcommand{\arraystretch}{0.7} \begin{tabular}{cccccc}
		\includegraphics[width=0.16\linewidth]{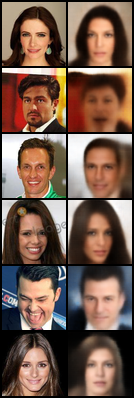} &
		\includegraphics[width=0.16\linewidth]{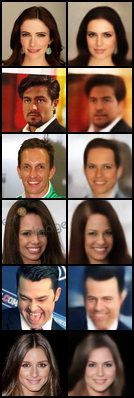} &
		\includegraphics[width=0.16\linewidth]{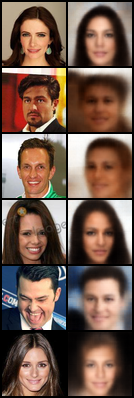}&
		\includegraphics[width=0.16\linewidth]{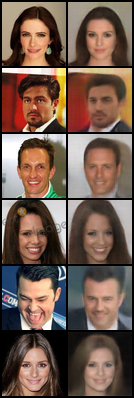} &
		\includegraphics[width=0.16\linewidth]{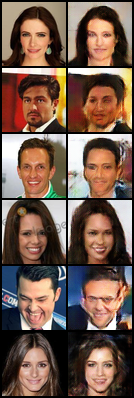} &
		\includegraphics[width=0.16\linewidth]{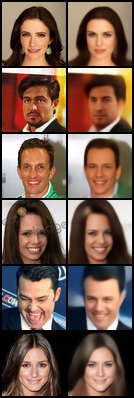}\\
	(a) VAE & (b) AAE & (c) WAE & (d) IntroVAE & (e) AGE & (f) Ours\\
		\\
	\end{tabular}
	\vspace{-5pt}
	\caption{
		The reconstructed results with different methods for CelebA Dataset.
	}
	\label{recon} \vspace{-10pt}
\end{figure*}

\section{Experiments} 
In this section, we evaluate our method on toy datasets and two widely-used
image datasets-MNIST~\cite{lecun1998mnist} and CelebA~\cite{liu2015deep}. As a preliminary evaluation,
we use low-dimensional datasets Swiss-roll to visually justify that our learned latent embedding can preserve the topological properties of the data manifold  and the prior can match our embedding distribution well, resulting in a satisfying reconstruction and generation simultaneously. Meanwhile, the two large scale datasets highlight a variety of challenges
that our method should address and evaluation on them is adequate to
support the advantages of ours.

\subsection{Swiss-roll}
We compare our method to the VAE~\cite{kingma_auto-encoding_2014}, AAE~\cite{2015Adversarial}, and ALAE~\cite{pidhorskyi2020Adversariala}. For each
method, we apply the same architectures with our paper. We uniformly sample 5000 samples on swiss-roll as the data manifold and use a 2D unit Gaussian distribution as our prior. Figure~\ref{recon_gene_swissroll} shows the reconstruction of the input data manifold and generated manifold with Gaussian samples as the input. Because of the reconstruction term in both VAE and AAE, they can get satisfying reconstructed manifolds. However, the generated manifolds are less satisfied due to the trade-off in their training. For ALAE, its principal training framework is based on GAN, not including a clear reconstruction loss. Therefore, ALAE fails to capture the topological structure of the original manifold even on generation task.  Figure~\ref{latent_representation} further illustrates the relation between latent representation and prior distribution. The latent embedding for VAE and AAE try to fit Gaussian distribution while preserving some geometric information but can't match Gaussian distribution absolutely. Due to the 
self drawbacks of GAN, the generated manifold using ALAE can not well preserve the geometry of data manifold. Although imposing reciprocity in the latent space gives some advantages to choosing measure metrics, they also can not preserve topological properties. Therefore, ALAE gets bad results for reconstruction. There is no trade-off problem with our approach. The latent embedding can fully preserve geometric information. With an adversarial strategy, we can obtain a satisfying latent distribution mapping to generate new data within original data manifold from a specified prior.

\subsection{Ablation Study} 
We do an ablation study on Mnist dataset to verify each component of our framework including adversarial regularizer and batch normalization(BN). The latent dimension is set to 256D. We show the reconstructed, interpolated, and generated results with different combinations of each component. We can observe that all of the comparison methods can get well-performed reconstruction results. However, without BN for latent embedding, we fail to generate digital images. This is mainly because the embedding distribution may be overdispersed in 256-dimensional latent space which is difficult for the prior to learn. This also brings more challenges to interpolation tasks. With BN operation, the generated images have better visual quality but may have some overlaps by two digits. After adding our adversarial regularizer, we can generate sharp digital images with high quality, as well as interpolated results. This demonstrates our adversarial regularizer can further improve the learning ability aimed at latent distribution while BN makes the embedding subjected to a distribution with its mean close to 0 and variance close to 1.

\subsection{CelebA Dataset}
For CelebA datasets, we measure the performance quantitatively with FID scores~\cite{heusel2017gans} for generated images and MSE metric for reconstructed images. FID can detect intra-class mode dropping, and measure the diversity as well as the quality of generated samples. All the training and testing images are resized to 64. FID is computed from 50K generated samples and the pre-calculated statistics are precomputed
on all training data. The model architecture for networks $G,g$  and $D,d$ follow the Mimicry's architectures~\cite{lee2020mimicry}. We compare our
model with other autoencoder based methods including VAE~\cite{kingma_auto-encoding_2014}, AAE~\cite{2015Adversarial}, WAE~\cite{Tolstikhin2017Wasserstein}, introVAE~\cite{2018IntroVAE} and AGE~\cite{ulyanov2017It}. Each model is trained for 25 epochs. Table~\ref{evaluation} shows the quantitative evaluation results. Our method obtains the best results both on FID  and MSE, indicating that our method can achieve better 
performance both on reconstruction and generation rather than a trade-off between them. Moreover, we show the qualitative visual results for our approach and those comparison methods above in Fig.~\ref{gene} and Fig.~\ref{recon}. We can observe our method produces visually appealing results both in reconstruction and sampling. Both the quantitative and qualitative results demonstrate our method is able to preserve the most global topology information of the input data manifold while achieving high-quality generation in visual perception. 
\begin{table}[]
	\centering
	\renewcommand{\arraystretch}{1.3}
	\resizebox{\linewidth}{!}{
	\begin{tabular}{c|cccccc}
		Methods & VAE & AAE & WAE & \multicolumn{1}{l}{IntroVAE} & AGE & Ours \\ \hline
		FID & 80 & 101 & 115 & 120 & 116 & 62 \\
		MSE & 0.0383  & \multicolumn{1}{l}{0.0255} & 0.0626 & 0.0562 & 0.0889 & 0.0165
	\end{tabular}}
	\caption{The FID and MSE values for different methods on CelebA Dataset. }
	\label{evaluation}
	%\hspace{-10pt}
\end{table}
% \usepackage{multirow}
\begin{table}[]
\centering
\renewcommand{\arraystretch}{1.3}
\resizebox{\linewidth}{!}{
\begin{tabular}{c|c|cccccl}
Metrics & \multicolumn{1}{l|}{Datasets} & VAE & AAE & WAE & \multicolumn{1}{l}{IntroVAE} & AGE & \multicolumn{1}{c}{Ours} \\ \hline
\multirow{2}{*}{FID} & cifar10 & 189 & 120 & 173 & 133 & 271 &  \\
 & CelebA & 80 & 101 & 115 & 120 & 116 & \multicolumn{1}{c}{63} \\ \hline
\multirow{2}{*}{MSE} & cifar10 & \multicolumn{1}{l}{0.0626} & \multicolumn{1}{l}{0.0245} & \multicolumn{1}{l}{0.0607} & 0.0492 & \multicolumn{1}{l}{0.0608} &  \\
 & CelebA & \multicolumn{1}{l}{0.0383} & \multicolumn{1}{l}{0.0255} & \multicolumn{1}{l}{0.0626} & 0.0562 & \multicolumn{1}{l}{0.0889} & 0.0165
\end{tabular}}
\caption{}
\label{tab:my-table}
\end{table}
%\begin{table}[t]
%\centering
%%\resizebox{.95\columnwidth}{!}{
%\begin{tabular}{l|l|l|l}
%    authblk & babel & cjk & dvips \\
%    epsf & epsfig & euler & float \\
%    fullpage & geometry & graphics & hyperref \\
%    layout & linespread & lmodern & maltepaper \\
%    navigator & pdfcomment & pgfplots & psfig \\
%    pstricks & t1enc & titlesec & tocbind \\
%    ulem
%\end{tabular}
%\caption{LaTeX style packages that must not be used.}
%\label{table2}
%\end{table}
\section{Conclusion}
Despite the recent success of variational autoencoder and its variants, there exists a trade-off between reconstruction and generation. Rather than imposing the latent distribution matching the prior, we propose to let the prior fit a learned latent representation which preserves the topological properties of the data manifold. 
In order to learn a latent distribution which is conducive to prior learning, 
we introduce an adversarial regularizer on interpolated samples and a batch normalization trick on latent embedding to obtain a concentrated latent distribution. 
We give a theoretical analysis of the superiority of our method and perform extensive experiments to further verify its effectiveness. Our experiments show that our method is able to preserve the most global topology information  of  the  input  data  manifold  while achieving high-quality generation without sampling meaningless points.

%\begin{figure}[t]
%\centering
%\includegraphics[width=0.9\columnwidth]{figure1} % Reduce the figure size so that it is slightly narrower than the column. Don't use precise values for figure width.This setup will avoid overfull boxes.
%\caption{Using the trim and clip commands produces fragile layers that can result in disasters (like this one from an actual paper) when the color space is corrected or the PDF combined with others for the final proceedings. Crop your figures properly in a graphics program -- not in LaTeX}.
%\label{fig1}
%\end{figure}
%
%\begin{figure*}[t]
%\centering
%\includegraphics[width=0.8\textwidth]{figure2} % Reduce the figure size so that it is slightly narrower than the column.
%\caption{Adjusting the bounding box instead of actually removing the unwanted data resulted multiple layers in this paper. It also needlessly increased the PDF size. In this case, the size of the unwanted layer doubled the paper's size, and produced the following surprising results in final production. Crop your figures properly in a graphics program. Don't just alter the bounding box.}
%\label{fig2}
%\end{figure*}

% Using the \centering command instead of \begin{center} ... \end{center} will save space
% Positioning your figure at the top of the page will save space and make the paper more readable
% Using 0.95\columnwidth in conjunction with the

%\section{ Acknowledgments}

\bibliographystyle{plain}
\bibliography{ref.bib}

\begin{thebibliography}{26}
\providecommand{\natexlab}[1]{#1}
\providecommand{\url}[1]{\texttt{#1}}
\providecommand{\urlprefix}{URL }
\expandafter\ifx\csname urlstyle\endcsname\relax
  \providecommand{\doi}[1]{doi:\discretionary{}{}{}#1}\else
  \providecommand{\doi}{doi:\discretionary{}{}{}\begingroup
  \urlstyle{rm}\Url}\fi

\bibitem[{Berthelot et~al.(2019)Berthelot, Raffel, Roy, and
  Goodfellow}]{berthelot2019Understandinga}
Berthelot, D.; Raffel, C.; Roy, A.; and Goodfellow, I. 2019.
\newblock Understanding and Improving Interpolation in Autoencoders via an
  Adversarial Regularizer.
\newblock In \emph{Proceedings of the 7th {{International Conference}} on
  {{Learning Representations}}}.

\bibitem[{Blum, Hopcroft, and Kannan(2016)}]{blum2016foundations}
Blum, A.; Hopcroft, J.; and Kannan, R. 2016.
\newblock Foundations of data science.
\newblock \emph{Vorabversion eines Lehrbuchs} 5.

\bibitem[{Bouchacourt, Tomioka, and Nowozin(2017)}]{Bouchacourt2017Multi}
Bouchacourt, D.; Tomioka, R.; and Nowozin, S. 2017.
\newblock Multi-Level Variational Autoencoder: Learning Disentangled
  Representations from Grouped Observations.
\newblock \emph{arXiv preprint arXiv:1705.08841} .

\bibitem[{Brock, Donahue, and Simonyan(2019)}]{brock2019Large}
Brock, A.; Donahue, J.; and Simonyan, K. 2019.
\newblock Large {{Scale GAN Training}} for {{High Fidelity Natural Image
  Synthesis}}.
\newblock \emph{Proceedings of the 2nd International Conference on Learning
  Representations} .

\bibitem[{Davidson et~al.(2018)Davidson, Falorsi, De~Cao, Kipf, and
  Tomczak}]{davidson2018hyperspherical}
Davidson, T.~R.; Falorsi, L.; De~Cao, N.; Kipf, T.; and Tomczak, J.~M. 2018.
\newblock Hyperspherical variational auto-encoders.
\newblock \emph{arXiv preprint arXiv:1804.00891} .

\bibitem[{Goodfellow et~al.(2014)Goodfellow, Pouget-Abadie, Mirza, Xu,
  Warde-Farley, Ozair, Courville, and Bengio}]{Goodfellow2014Generative}
Goodfellow, I.~J.; Pouget-Abadie, J.; Mirza, M.; Xu, B.; Warde-Farley, D.;
  Ozair, S.; Courville, A.; and Bengio, Y. 2014.
\newblock Generative Adversarial Networks.
\newblock \emph{Advances in Neural Information Processing Systems} 3:
  2672--2680.

\bibitem[{Heusel et~al.(2017)Heusel, Ramsauer, Unterthiner, Nessler, and
  Hochreiter}]{heusel2017gans}
Heusel, M.; Ramsauer, H.; Unterthiner, T.; Nessler, B.; and Hochreiter, S.
  2017.
\newblock Gans trained by a two time-scale update rule converge to a local nash
  equilibrium.
\newblock In \emph{Advances in neural information processing systems},
  6626--6637.

\bibitem[{Huang et~al.(2018)Huang, Li, He, Sun, and Tan}]{2018IntroVAE}
Huang, H.; Li, Z.; He, R.; Sun, Z.; and Tan, T. 2018.
\newblock IntroVAE: Introspective Variational Autoencoders for Photographic
  Image Synthesis.
\newblock \emph{Advances in Neural Information Processing Systems} .

\bibitem[{Kingma and Welling(2014)}]{kingma_auto-encoding_2014}
Kingma, D.~P.; and Welling, M. 2014.
\newblock Auto-Encoding Variational Bayes.
\newblock In \emph{Proceedings of the 2nd International Conference on Learning
  Representations}.

\bibitem[{Larsen and Winther(2016)}]{vaegan}
Larsen, Anders Boesen~Lindbo, S. K. S. H.~L.; and Winther, O. 2016.
\newblock Autoencoding beyond pixels using a learned similarity metric.
\newblock \emph{International conference on machine learning} .

\bibitem[{LeCun(1998)}]{lecun1998mnist}
LeCun, Y. 1998.
\newblock The MNIST database of handwritten digits.
\newblock \emph{http://yann. lecun. com/exdb/mnist/} .

\bibitem[{Lee and Town(2020)}]{lee2020mimicry}
Lee, K.~S.; and Town, C. 2020.
\newblock Mimicry: Towards the Reproducibility of GAN Research .

\bibitem[{Liu et~al.(2015)Liu, Luo, Wang, and Tang}]{liu2015deep}
Liu, Z.; Luo, P.; Wang, X.; and Tang, X. 2015.
\newblock Deep learning face attributes in the wild.
\newblock In \emph{Proceedings of the IEEE international conference on computer
  vision}, 3730--3738.

\bibitem[{Makhzani et~al.(2015)Makhzani, Shlens, Jaitly, and
  Goodfellow}]{2015Adversarial}
Makhzani, A.; Shlens, J.; Jaitly, N.; and Goodfellow, I. 2015.
\newblock Adversarial Autoencoders.
\newblock \emph{arXiv preprint arXiv:1511.05644} .

\bibitem[{Mathieu and Teh(2019)}]{disentanglement}
Mathieu, Emile, T. R. N.~S.; and Teh, Y.~W. 2019.
\newblock Disentangling disentanglement in variational autoencoders.
\newblock \emph{International Conference on Machine Learning} .

\bibitem[{Oord, Vinyals, and Kavukcuoglu(2017)}]{2017Neural}
Oord, A. V.~D.; Vinyals, O.; and Kavukcuoglu, K. 2017.
\newblock Neural Discrete Representation Learning.
\newblock \emph{Advances in Neural Information Processing Systems} .

\bibitem[{Pidhorskyi, Adjeroh, and Doretto(2020)}]{pidhorskyi2020Adversariala}
Pidhorskyi, S.; Adjeroh, D.~A.; and Doretto, G. 2020.
\newblock Adversarial {{Latent Autoencoders}}.
\newblock In \emph{2020 {{IEEE}}/{{CVF Conference}} on {{Computer Vision}} and
  {{Pattern Recognition}}}. {Seattle, WA, USA}.
\newblock ISBN 978-1-72817-168-5.
\newblock \doi{10.1109/CVPR42600.2020.01411}.

\bibitem[{Qian et~al.(2019)Qian, Li, Cao, Liu, and Wong}]{2019Improving}
Qian, S.; Li, G.; Cao, W.~M.; Liu, C.; and Wong, H.~S. 2019.
\newblock Improving representation learning in autoencoders via
  multidimensional interpolation and dual regularizations.
\newblock In \emph{Twenty-Eighth International Joint Conference on Artificial
  Intelligence}.

\bibitem[{Tolstikhin et~al.(2018)Tolstikhin, Bousquet, Gelly, and
  Schoelkopf}]{Tolstikhin2017Wasserstein}
Tolstikhin, I.; Bousquet, O.; Gelly, S.; and Schoelkopf, B. 2018.
\newblock Wasserstein Auto-Encoders.
\newblock \emph{Proceedings of the 2nd International Conference on Learning
  Representations} .

\bibitem[{Ulyanov, Vedaldi, and Lempitsky(2017)}]{ulyanov2017It}
Ulyanov, D.; Vedaldi, A.; and Lempitsky, V. 2017.
\newblock It {{Takes}} ({{Only}}) {{Two}}: {{Adversarial Generator}}-{{Encoder
  Networks}}.
\newblock \emph{Proceedings of the AAAI Conference on Artificial Intelligence}
  .

\bibitem[{van Handel(2014)}]{van2014probability}
van Handel, R. 2014.
\newblock Probability in high dimension.
\newblock Technical report, PRINCETON UNIV NJ.

\bibitem[{Vincent et~al.(2010)Vincent, Larochelle, Lajoie, Bengio, and
  Manzagol}]{2010Stacked}
Vincent, P.; Larochelle, H.; Lajoie, I.; Bengio, Y.; and Manzagol, P.~A. 2010.
\newblock Stacked Denoising Autoencoders: Learning Useful Representations in a
  Deep Network with a Local Denoising Criterion.
\newblock \emph{Journal of Machine Learning Research} 11(12): 3371--3408.

\bibitem[{Xu and Durrett(2018)}]{2018Spherical}
Xu, J.; and Durrett, G. 2018.
\newblock Spherical Latent Spaces for Stable Variational Autoencoders.
\newblock \emph{Conference on Empirical Methods in Natural Language Processing}
  .

\bibitem[{Zhang et~al.(2019)Zhang, Goodfellow, Metaxas, and
  Odena}]{zhang2019SelfAttention}
Zhang, H.; Goodfellow, I.; Metaxas, D.; and Odena, A. 2019.
\newblock Self-{{Attention Generative Adversarial Networks}}.
\newblock In \emph{International {{Conference}} on {{Machine Learning}}}.

\bibitem[{Zhao, Zhu, and Zhang(2019)}]{zhao2019latent}
Zhao, D.; Zhu, J.; and Zhang, B. 2019.
\newblock Latent Variables on Spheres for Autoencoders in High Dimensions.
\newblock \emph{arXiv} arXiv--1912.

\bibitem[{Zhao, Song, and Ermon(2019)}]{zhao2019infovae}
Zhao, S.; Song, J.; and Ermon, S. 2019.
\newblock Infovae: Balancing learning and inference in variational
  autoencoders.
\newblock In \emph{Proceedings of the aaai conference on artificial
  intelligence}, volume~33, 5885--5892.

\end{thebibliography}

\end{document}